%% file: iclr2025_conference.tex
\documentclass{article} 
\usepackage{iclr2025_conference,times}
\usepackage{multirow}
\usepackage{graphicx}
\usepackage{booktabs}
\usepackage{amssymb}
\usepackage{amsmath}
\usepackage{algorithm}
\usepackage{algorithmicx}
\usepackage{algpseudocode}
\input{math_commands.tex}

\usepackage{hyperref}
\usepackage{url}

\title{Detection Limits and Statistical Separability of Tree Ring Watermarks in Rectified Flow-based Text-to-Image Generation Models}

\author{Ved Umrajkar\thanks{Equal contribution. Data Science Group, IIT Roorkee.} \\ 
Department of Mathematics\\
Indian Institute of Technology Roorkee\\
\texttt{v\_umrajkar@ma.iitr.ac.in} \\
\And 
Aakash Kumar Singh\footnotemark[1] \\ 
Mehta Family School of Data Science and Artificial Intelligence \\
Indian Institute of Technology Roorkee \\
\texttt{aakash\_ks@mfs.iitr.ac.in}
}

\iclrfinalcopy

\begin{document}
\maketitle

\begin{abstract} 
Tree-Ring Watermarking is a significant technique for authenticating AI-generated images. However, its effectiveness in rectified flow-based models remains unexplored, particularly given the inherent challenges of these models with noise latent inversion. Through extensive experimentation, we evaluated and compared the detection and separability of watermarks between SD 2.1 and FLUX.1-dev models. By analyzing various text guidance configurations and augmentation attacks, we demonstrate how inversion limitations affect both watermark recovery and the statistical separation between watermarked and unwatermarked images. Our findings provide valuable insights into the current limitations of Tree-Ring Watermarking in the current SOTA models and highlight the critical need for improved inversion methods to achieve reliable watermark detection and separability. The official implementation, dataset release and all experimental results are available at this \href{https://github.com/dsgiitr/flux-watermarking}{\textbf{link}}.
\end{abstract}

\section{Introduction}

The rapid advancement of generative AI models has raised pressing concerns about the authenticity and provenance of digital content. While watermarking techniques for AI-generated images have emerged as a promising solution, their effectiveness heavily depends on reliable detection and clear separability between watermarked and non-watermarked content. Recent approaches like Tree Ring Watermarking \citep{wen2024tree} have shown promise, but their effectiveness remains unexplored for newer architectures.

Recent advances in text-conditioned generative models, particularly rectified flow models, have demonstrated remarkable capabilities in high-resolution image synthesis. Unlike traditional diffusion models, rectified flows model transportation between distributions through linear interpolation of marginals, enabling efficient sampling with fewer discretization steps. However, the implications of these architectural differences on watermarking mechanisms remain unexplored.

This work investigates watermark detection and separability in flow-based generative models, focusing on the \cite{FLUX} model. We analyze two critical aspects: the reconstruction and detection of embedded watermarks through noise latent inversion, and the statistical separability between watermarked and non-watermarked distributions under various attack scenarios. Our findings demonstrate that while flow-based models present unique challenges for watermark detection, careful consideration of model configuration and inversion methodology can achieve reliable separation.

\section{Related Works}

\paragraph{Watermarking Approaches.}

Random seed modification watermarks like Tree Ring~\citep{wen2024tree} and RingID~\citep{ci2024ringid} embed a known key into the noise latent that is the starting point for image generation using diffusion. The effectiveness of these approaches has been systematically evaluated through benchmarks like Waves \citep{anwaves}, which provides standardized attack scenarios for robustness assessment.

\paragraph{Inversion Methods.}
Recent work has advanced latent inversion techniques, with \cite{hong2024exact} demonstrating significantly improved Tree-ring watermark detection using higher-order inversion algorithms compared to naive DDIM inversion. While their work showed remarkable detection accuracy on traditional diffusion models using DPM-Solver++~\citep{lu2022dpmpp}, the effectiveness of these techniques on newer rectified flow-based architectures remains unexplored. Our work extends this analysis to flow-based models, providing insights into watermark detection across different architectures.




\section{Methodology}
\begin{figure}
    \centering
    \includegraphics[width=1.1\linewidth]{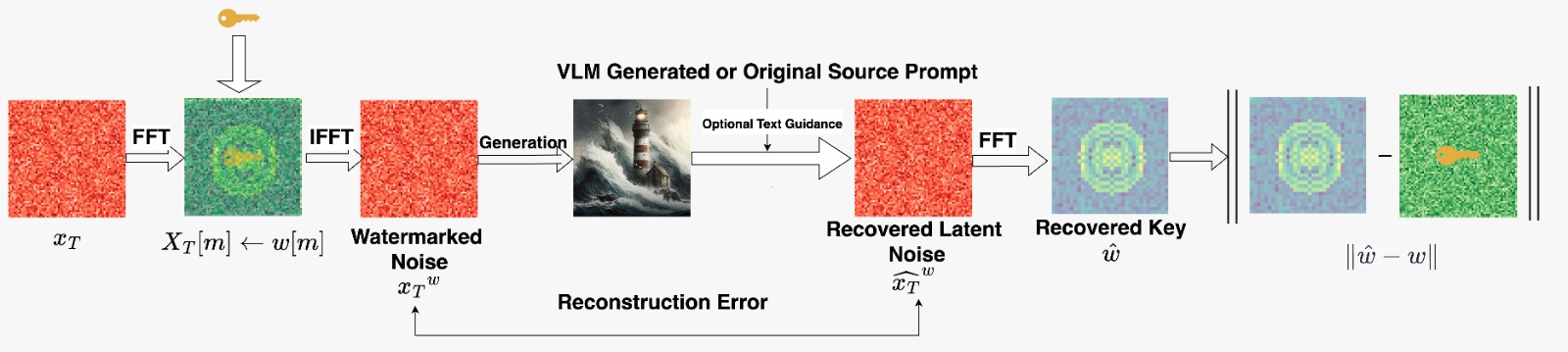}
    \caption{Watermarking workflow for both FLUX and Stable Diffusion}
    \label{fig:enter-label}
\end{figure}

\subsection{Preliminaries}

\paragraph{Notation}

Let $x_t$ denote the noisy image at timestep $t$, with $x_0$ and $x_T$ representing the generated image and initial noise latent respectively. In frequency domain, $X_T$ denotes the Fourier transform of $x_T$. The data and noise distributions are denoted by $\pi_0$ and $\pi_1 \sim \mathcal{N}(0,I)$ respectively. We denote the parameters of a neural network by $\theta$ that can be used for adequate prediction targets and $c$ for the text prompt used to guide the text-to-image generation models.



\paragraph{Generation and Inversion Framework}
Image generation involves producing an image $x_0$ from random noise $x_T$, while inversion aims to reconstruct the original noise latent ($\hat{x}_T$) from an input generated latent. The noise map obtained from inversion should generate the exact same image $x_0$ by sampling using diffusion. Both processes involve solving Ordinary Differential Equations (ODEs) through numerical integration, which can be done using first-order methods, such as Euler's method. This is usually the case with models like Flux, which follow a more linear trajectory. We focus on two primary approaches: traditional Denoising Diffusion Models (DDMs) and the newer Rectified Flow Transformer models \citep{liu2023flow,lipman2022flow,liu2022rectifiedflow,esser2024scalingsd3}.


\paragraph{FLUX}
Transformers trained with the flow-matching objective have recently achieved state-of-the-art results in image generation \citep{esser2024scalingsd3}. We utilize the open weights \cite{2025bflFLUX1dev} model, which employs a Diffusion Transformer (DiT) \citep{peebles2023scalable} architecture and differs fundamentally from traditional DDMs like Stable Diffusion \citep{rombach2022sd2} in its approach to generation and inversion. FLUX is based on rectified flows, which construct a transportation between the source distribution $\pi_1$ (typically standard Gaussian) and the target data distribution $\pi_0$ through the following ODE:
\[
\frac{dx_t}{dt} = v_t(x_t,t,c)dt, \quad X_0 \sim \pi_0, \quad t \in [0,1]
\]
where $v_t$ is a time-dependent velocity field parameterized by the neural network. A key property of rectified flows is that the marginal distribution at time $t$ follows a linear interpolation between $x_0$ and $x_1$:
\[
x_t \sim (1-t)x_0 + tx_1
\]
This property enables efficient sampling with relatively few discretization steps. For generation, the ODE is solved forward, while inversion uses the backward Euler method:
\[
\mathbf{x}_{t_{i}} = \mathbf{x}_{t_{i-1}} - (t_{i} - t_{i-1})\,\mathbf{v}_\theta(\mathbf{x}_{t_i}, t_i, c)
\]
This contrasts with the usual first-order (naive) DDIM inversion:
\[
x_{t+1} = \sqrt{\bar{\alpha}_{t+1}}\,\hat{x}_0^t + \sqrt{1-\bar{\alpha}_{t+1}}\,\epsilon_\theta(x_t, \sigma_t, c)
\]

For more details and a thorough discussion, refer \hyperref[app:diffusion_flow]{Appendix Section A}.

\subsection{Approach}

\paragraph{Tree-Ring Watermark Embedding}
The Tree-Ring watermark embedding follows a Fourier space modification approach:
\begin{equation}
\mathbf{x}_T = \mathcal{F}^{-1}(\mathbf{X}_T), \quad \text{where } \mathbf{X}_T[m] = w[m]
\end{equation}
Here, $w$ represents the ring-pattern watermark key, $m$ the circular mask in channel $\mathcal{C}_w$, and $\mathcal{F}^{-1}$ the inverse Fourier transform. The corresponding recovered key is denoted by $\hat{w}$ which is obtained from the fourier transform of the recovered watermarked noise latent, i.e. \(\hat{w}=\mathcal{F}({\widehat{x_T}^w})\).

The complete watermarking procedure is detailed in Algorithm \ref{alg:watermark}.

\paragraph{VLM Generated Prompt for Inversion Guidance}
For real-world scenarios where original prompts might be unavailable, we employ Qwen2-VL-2B-Instruct~\citep{qwen2vl} to generate image-grounded captions as alternative prompts. This approach enables evaluation of both prompt-free and prompt-guided inversion scenarios.



\paragraph{Evaluation}
We evaluate watermark separability by analyzing the distribution of Fourier space distances $d = \|\hat{w} - w\|$ between reconstructed ($\hat{w}$) and original ($w$) watermark patterns. To quantify the statistical separation between different configurations (with/without prompts, with/without attacks), we compute the Symmetric KL Divergence between their respective distance distributions: $\mathcal{D}_{\text{SKL}}(P\|Q) = \frac{1}{2}[\mathcal{D}_{\text{KL}}(P\|Q) + \mathcal{D}_{\text{KL}}(Q\|P)]$, where P and Q represent the distance distributions for different experimental configurations.

\begin{algorithm}[t]
\caption{Tree Ring Watermarking Procedure}
\label{alg:watermark}
\begin{algorithmic}[1]
\Require Image dimensions $(h, w)$, watermark channel $c_w$, radius $r$, batch size $b$, seed $s$
\Ensure Watermarked noise $x_T$, watermark key $w$, watermark mask $m$

\State $\boldsymbol{x_T} \sim \mathcal{N}(0, I)$ \hfill \textit{(Sample initial Gaussian noise)}
\State Generate watermark mask $m$ using radius $r$ and channel $c_w$
\State Generate watermark key $w$ using pattern and seed $s$
\State Compute FFT of noise: $ X_T\gets \text{FFT}(\boldsymbol{x_T})$
\State Apply watermark: $\hat{X_T}[m] \gets w[m]$
\State Compute inverse FFT: $x_T^w \gets \text{IFFT}(\hat{X_T})$

\Return $x_T^w $, $w$, $m$
\end{algorithmic}
\end{algorithm}

\section{Experiments}

\subsection{Experimental Setup}

To ensure consistency, we used a fixed global random seed for generating initial latents, enabling reproducibility across models and configurations. The same watermark key, derived from this seed, was applied throughout the study. Additionally, we adopted a uniform timestep schedule for both sampling and inversion, which appeared to improve the inversion accuracy. We perform the experiments on the test partition of the open source Stable Diffusion Prompts Dataset \citep{gustavo2023sdprompts}. For all our experiments we have used a fixed Classifier-free guidance~\citep{bansalUniversalGuidanceDiffusion2023} of 3.5.

\subsection{Results and Analysis}

\begin{table}
\centering
\renewcommand{\arraystretch}{1.3}
\setlength{\tabcolsep}{3pt} 
\caption{Watermark Extraction Metrics}
\label{tab:watermark-metrics}
\begin{tabular}{llcccc}
\toprule
\textbf{Model} & \textbf{Configuration} & \multicolumn{2}{c}{Latent Noise Reconstruction} & $\frac{1}{n}\sum{|\hat{w_i} - w_i|}$ \\
\cmidrule(lr){3-4}
 &  & \textbf{NMAE} $\frac{|\hat{w} - w|_1}{|w|_1}$ & \textbf{NMSE} $\frac{|\hat{w} - w|_2^2}{|w|_2^2}$ &  \\
\midrule
\multirow{5}{*}{FLUX.1-dev} & No Attack (No Prompt) & $0.303_{0.056}$ & $0.106_{0.042}$ & $22.772_{0.550}$ \\
& No Attack (With Prompt) & $0.232_{0.046}$ & $0.063_{0.029}$ & $22.123_{0.351}$ \\
& No Attack (With VLM) & $0.290_{0.050}$ & $0.096_{0.037}$ &  $22.701_{0.477}$ \\
\cmidrule{2-5}
& Blur (No Prompt) & $\mathbf{1.259_{0.016}}$ & $\mathbf{1.594_{0.040}}$ & $37.514_{1.066}$ \\
& Blur (With Prompt) & $\mathbf{1.261_{0.017}}$ & $\mathbf{1.597_{0.042}}$ & $37.718_{1.049}$ \\
& Noise (No Prompt) & $\mathbf{1.325_{0.043}}$ & $\mathbf{1.772_{0.018}}$ & $38.656_{1.372}$ \\
& Noise (With Prompt) & $\mathbf{1.343_{0.043}}$ & $\mathbf{1.821_{0.106}}$ & $39.309_{1.302}$ \\
\midrule
\multirow{2}{*}{SD 2.1 base} & No Attack (No Prompt) & $0.345_{0.060}$ & $0.132_{0.047}$ & $45.601_{1.702}$ \\
& No Attack (With Prompt) & $0.338_{0.061}$ & $0.128_{0.047}$ & $45.070_{1.749}$ \\
\bottomrule
\multicolumn{5}{l}{\small Note: Subscripts denote standard deviations. Last column represents average L1 distance in Fourier space.} \\
\end{tabular}
\end{table}

\begin{figure}[t]
    \centering
    \includegraphics[width=\textwidth]{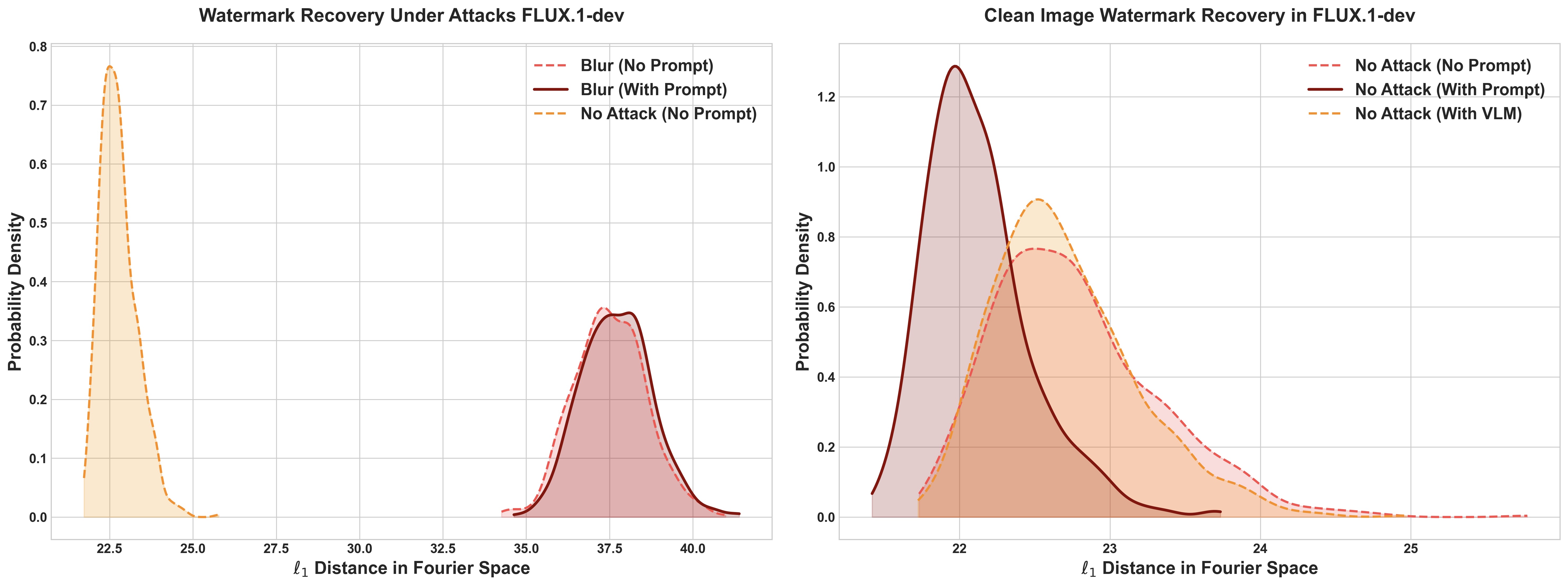}
    \caption{Distribution of watermark distances in Fourier space. Attacked scenarios show the distribution of the fourier space distance under noise, and blur manipulations. It can be clearly seen that in non-attacked scenarios, the prompt guidance plays a significant role in accurate inversion. We note that in attack scenarios the distance in the fourier space is drastically increased for FLUX.1-dev. }
    \label{fig:watermark-dist}
\end{figure}

\begin{table}
\centering
\caption{AUC comparison for watermark detection under different attacks}
\label{table:main_attack}
\begin{tabular}{cccc}
\toprule
Model & Blurring & Noise \\ \midrule
\textit{SD 2.1 base (DDIM)} & 0.999 & 0.944  \\
\textit{FLUX.1-dev (RF)} & \textbf{0.888} & \textbf{0.662} \\
\bottomrule
\end{tabular}
\vspace{-.25cm}
\end{table}

\paragraph{Clean Images.} Our experiments with non-attacked scenarios reveal that exact prompt guidance during inversion yields the lowest reconstruction error in both Fourier and spatial domains however, in case of attacked images the exact prompt guidance does not aid in reconstruction. Interestingly, FLUX.1-dev demonstrates superior latent noise reconstruction for clean images compared to the baseline model SD 2.1 \citep{rombach2022sd2}. However, this advantage diminishes significantly under attacked scenarios, where the separability between watermarked and non-watermarked distributions becomes drastically reduced.

The use of VLM-generated prompts demonstrates a noteworthy but constrained improvement in watermark detection. While these semantically derived prompts show marginal benefits in distribution separability and offer performance intermediate between exact prompt and no-prompt configurations, they fail to match the effectiveness of exact prompt guidance. This suggests that while semantic understanding from VLMs can aid reconstruction, \textbf{precise prompt matching remains crucial for optimal watermark recovery}.

DDIM (SD 2.1 base) exhibits robust separation between watermarked and non-watermarked images through naive inversion, maintaining consistent performance regardless of prompt guidance. This behavior contrasts significantly with FLUX.1-dev, where reconstruction quality demonstrates marked sensitivity to the presence and accuracy of prompt guidance.

\paragraph{Attacked Scenarios.} Under attacked scenarios, the application of noise and blur perturbations significantly compromises the watermark detection capability in FLUX.1-dev as shown in \ref{tab:watermark-metrics}. This degradation is particularly evident in the Fourier domain, where the characteristic ring patterns become increasingly difficult to distinguish from background frequencies. This behavior stands in stark contrast with DDIM, where previous work by\citep{wen2024tree}has demonstrated that latent noise reconstruction maintains high fidelity even under various attack scenarios.


\begin{table}[t]
\caption{Watermark Detection Performance for clean images with FLUX.1-dev}
\label{tab:watermark-detection-clean}
\centering
\begin{tabular}{lccccc}
\toprule
Configuration & TPR@1\%FPR & AUC & \multicolumn{3}{c}{Thresholds at FPR} \\
\cmidrule(lr){4-6}
& & & 1\% & 5\% & 10\% \\
\midrule
No Prompt & 1.000 & 0.989 & 36.950 & 37.538 & 37.978 \\
With Prompt & 1.000 & 0.999 & 36.949 & 37.524 & 37.993 \\
\bottomrule
\multicolumn{6}{l}{}
\end{tabular}
\end{table}

\paragraph{Impediments to Watermark Recovery}
The observed differences in watermark recovery between FLUX.1-dev and traditional diffusion models stem from fundamental architectural and training methodology differences. Flux employs a Multimodal Diffusion Transformer (MM DiT) architecture where text and image information are deeply entangled throughout the network, unlike older diffusion models' UNet architecture where text conditioning occurs primarily through cross-attention layers. This architectural difference makes image generation in Flux more fundamentally dependent on prompt information. Additionally, Flux uses a T5 text encoder with different latent characteristics than the CLIP encoder used in stable diffusion models, further altering information flow through the model. Most importantly, the rectified flow training objective optimizes for straight paths between source and target distributions, prioritizing efficient forward sampling at the expense of invertibility. This straightened path inherently discards information that would be useful during inversion. Higher-order numerical methods might offer incremental improvements, but cannot fully overcome these fundamental architectural limitations.

\section{Conclusion and Future Work}

Our study reveals fundamental differences in watermark detection and recovery capabilities across DDIM (SD 2.1 base) and FLUX.1-dev architectures. Most notably, we find that the diffusion transformer model FLUX.1-dev exhibits a strong dependency on prompt guidance for accurate reconstruction and watermark recovery, differing significantly from DDIM-based models like Stable Diffusion 2.1, which achieve reliable separation between watermarked and non-watermarked images even without prompt guidance and under attacks. Our analysis demonstrates that detection accuracy in FLUX.1-dev degrades significantly under attacked scenarios, underscoring the need for more robust inversion techniques. A qualitative visualization of image reconstruction is provided in Figure \ref{fig:noise_reconstruction}.

These findings highlight several critical directions for future research: developing improved inversion techniques specifically for rectified flow-based generative models, and crafting approaches to increase robustness of popular watermarking techniques over image manipulation attacks while maintaining watermark effectiveness.

\clearpage

\bibliography{iclr2025_conference}
\bibliographystyle{iclr2025_conference}

\input{appendix}

\end{document}

%% file: math_commands.tex
\usepackage{amsmath,amsfonts,bm}

\def\eqref#1{equation~\ref{#1}}

\def\1{\bm{1}}

\DeclareMathAlphabet{\mathsfit}{\encodingdefault}{\sfdefault}{m}{sl}
\SetMathAlphabet{\mathsfit}{bold}{\encodingdefault}{\sfdefault}{bx}{n}



%% file: appendix.tex
\appendix
\section{Appendix: Diffusion Models and Flow Matching}
\label{app:diffusion_flow}

\subsection{Latent Diffusion Models}
\label{subsec:ldm}

Latent Diffusion Models \citep{rombach2022high} (LDMs) operate in the compressed latent space of an autoencoder rather than directly in pixel space. The autoencoder consists of an encoder $\mathcal{E}$ that maps images $x \in \mathbb{R}^{H \times W \times 3}$ to a lower-dimensional latent representation $z = \mathcal{E}(x) \in \mathbb{R}^{h \times w \times c}$, and a decoder $\mathcal{D}$ that reconstructs the image from latents.

The diffusion process occurs entirely in this latent space, offering two key advantages: reduced computational complexity due to lower dimensionality, and the ability to leverage semantic compression from the autoencoder. Given a noise schedule $\{\beta_t\}_{t=1}^T$ and defining $\bar{\alpha}_t = \prod_{i=1}^t (1-\beta_i)$, the forward process adds noise to the latents:

\begin{equation}
   z_t = \sqrt{\bar{\alpha}_t}z_0 + \sqrt{1-\bar{\alpha}_t}\epsilon, \quad \epsilon \sim \mathcal{N}(0, I)
\end{equation}

where $z_0 = \mathcal{E}(x)$ is the encoded latent. The model learns to predict the noise component using a neural network $\epsilon_\theta(z_t, t)$ trained with the objective:

\begin{equation}
   \mathcal{L}_\text{simple} = \mathbb{E}_{t,z_0,\epsilon}\left[\|\epsilon - \epsilon_\theta(z_t,t)\|_2^2\right]
\end{equation}

After the diffusion and denoising process, the final latent $z_0$ is decoded to obtain the image: $x = \mathcal{D}(z_0)$. The LDM architecture's latent space dimensions vary across implementations. The FLUX dev model uses a VAE with latent dimensions $(16, h/8, w/8)$ where $h,w$ are the input image dimensions, allowing for flexible resolution generation. In contrast, Stable Diffusion 2.1 base model employs a fixed latent dimension of $(4, 64, 64)$

\subsection{DDIM Sampling and Inversion}
\label{subsec:ddim}

Denoising Diffusion Implicit Models (DDIM) provide a deterministic framework for generating images through the reverse diffusion process. Unlike standard diffusion models, DDIM defines a non-Markovian reverse process that enables deterministic trajectories between noise and images.

\begin{equation}
    x_{t-1} = \sqrt{\bar{\alpha}_{t-1}}\hat{x}_0^t + \sqrt{1-\bar{\alpha}_{t-1}}\epsilon_\theta(x_t,t)
\end{equation}

where $\hat{x}_0^t$ represents the predicted clean image:

\begin{equation}
    \hat{x}_0^t = \frac{x_t - \sqrt{1-\bar{\alpha}_t}\epsilon_\theta(x_t,t)}{\sqrt{\bar{\alpha}_t}}
\end{equation}

For inversion, DDIM maps a given image $x_0$ back to noise $x_T$ using:

\begin{equation}
    x_{t+1} = \sqrt{\bar{\alpha}_{t+1}}\hat{x}_0^t + \sqrt{1-\bar{\alpha}_{t+1}}\epsilon_\theta(x_t,t)
\end{equation}

This naïve DDIM inversion can be interpreted as forward Euler integration starting from $t=0$. While computationally efficient, it can accumulate errors over multiple steps.

\subsection{Rectified Flow and Flow Matching}
\label{subsec:rf}

Rectified Flow (RF) facilitates the transition between the data distribution $\pi_{0}$ and Gaussian noise distribution $\pi_{1}$ along a straight path. This is achieved by learning a forward-simulating system defined by the ODE:

\begin{equation}
    d\mathbf{x}_t = \mathbf{v}_\theta(\mathbf{x}_t,t)dt, \quad t \in [0,1]
\end{equation}

which maps $\mathbf{x}_1 \sim \pi_1$ to $\mathbf{x}_0 \sim \pi_0$. In practice, the velocity field $\mathbf{v}$ is parameterized by a neural network with parameters $\theta$.

During training, given empirical observations of two distributions $\mathbf{x}_0 \sim \pi_0$, $\mathbf{x}_1 \sim \pi_1$ and $t \in [0,1]$, the forward process of rectified flow is defined by a simple linear combination:

\begin{equation}
    \mathbf{x}_t = t\mathbf{x}_1 + (1-t)\mathbf{x}_0
\end{equation}

which can be written in differential form as:
\begin{equation}
\label{eqn:ode}
    d\mathbf{x}_t = (\mathbf{x}_1 - \mathbf{x}_0)dt
\end{equation}

Consequently, the training process optimizes the network by solving the least squares regression problem:

\begin{equation}
    \min_\theta \int_0^1 \mathbb{E}\left[\|(\mathbf{x}_1 - \mathbf{x}_0) - \mathbf{v}_\theta(\mathbf{x}_t,t)\|^2\right] dt
\end{equation}

For sampling, the ODE \eqref{eqn:ode} is discretized and solved using the Euler method. The model starts with a Gaussian noise sample $\mathbf{x}_{t_N} \sim \mathcal{N}(0,I)$. Given a series of $N$ discrete timesteps $t = \{t_N,\ldots,t_0\}$, the model iteratively applies:

\begin{equation}
    \mathbf{x}_{t_{i-1}} = \mathbf{x}_{t_i} + (t_{i-1} - t_i)\mathbf{v}_\theta(\mathbf{x}_{t_i},t_i)
\end{equation}

For inversion, the backward Euler method is used:
\begin{equation}
    \mathbf{x}_{t_{i}} = \mathbf{x}_{t_{i-1}} - (t_{i} - t_{i-1})\mathbf{v}_\theta(\mathbf{x}_{t_i}, t_i)
\end{equation}

The RF model can generate high-quality images in much fewer timesteps compared to DDPM, owing to the nearly linear transition trajectory established during training.

\subsection{Higher-Order Inversion Methods}
\label{subsec:higher_order}

Recent work has introduced exact inversion techniques using higher-order ODE solvers. For DDIM, the backward Euler method provides more accurate inversion by solving:

\begin{equation}
    \hat{z}_{t_{i-1}} = \hat{z}_{t_i} - (t_i-t_{i-1})v_\theta(\hat{z}_{t_i},t_i)
\end{equation}

This can be improved through gradient descent steps:

\begin{equation}
    \nabla_{\hat{z}_{t_{i-1}}} \|\hat{z}_{t_i} - z'_{t_i}\|^2
\end{equation}

where $z'_{t_i}$ is computed using:

\begin{equation}
    z'_{t_i} \leftarrow \frac{\sigma_{t_i}}{\sigma_{t_{i-1}}}\hat{z}_{t_{i-1}} - \alpha_{t_i}(e^{-h_i}-1)z_0(\hat{z}_{t_{i-1}},t_{i-1})
\end{equation}

The DPM-Solver++ framework generalizes this to higher orders using the exponential integrator:

\begin{equation}
\begin{split}
    x_{t_i} = \frac{\sigma_{t_i}}{\sigma_{t_{i-1}}}x_{t_{i-1}} + \sigma_{t_i}\sum_{n=0}^{k-1}x_\theta^{(n)}(x_{\lambda_{t_{i-1}}},\lambda_{t_{i-1}}) \\
    \cdot\int_{\lambda_{t_{i-1}}}^{\lambda_{t_i}}\frac{e^\lambda(\lambda-\lambda_{t_{i-1}})^n}{n!}d\lambda
\end{split}
\end{equation}

where $\lambda_t = \log(\alpha_t/\sigma_t)$ is the log-SNR and $k$ represents the order of the solver.

\section{Appendix: Experimental Details}
\label{app:experiment}

\subsection{Model Configurations and Sampling Parameters}

We conducted our experiments using carefully calibrated configurations for both FLUX.1-dev and Stable Diffusion 2.1 models. The key parameters were selected to balance generation quality with computational efficiency while maintaining fair comparison conditions across models.

\subsubsection{FLUX-dev Configuration}
For the FLUX-dev model, we employed the following parameters:
\begin{itemize}
    \item Number of sampling steps: 28 steps for both generation and inversion processes
    \item Guidance scale: 3.5 (classifier-free guidance)
    \item Sampling method: Euler solver for ODE integration
    \item Timestep scheduling: Uniform spacing between t=0 and t=1
\end{itemize}

The relatively lower number of steps (28) for FLUX-dev is justified by its efficient rectified flow training objective and Euler integration scheme, which allows for larger step sizes while maintaining generation quality.

\subsubsection{Stable Diffusion 2.1 Configuration}
For SD2.1 with DDIM sampling, we used:
\begin{itemize}
    \item Number of sampling steps: 50 steps for both generation and inversion processes
    \item Guidance scale: 3.5 (matching FLUX-dev for comparative analysis)
    \item Sampling method: DDIM deterministic sampling
    \item Timestep scheduling: Default DDIM schedule
\end{itemize}

The higher number of steps (50) for DDIM sampling is necessary for finer granularity in the diffusion process timestep discretization.




These configurations were held constant across all experiments to ensure consistency and reproducibility of our results. The parameters were validated through preliminary experiments to ensure they produced high-quality generations while maintaining reasonable computational requirements.

\subsubsection{Evaluation Framework}
We quantify watermark robustness through the following metrics:
\begin{itemize}
    \item Fourier Space $\mathcal{L}_1$ Distance: Measures discrepancy between reconstructed ($\hat{w}$) and original ($w$) watermark patterns in frequency domain:
    \[
        \mathcal{L}_1(w, \hat{w}) = \sum_{i} |w_i - \hat{w}_i|
    \]
    \item Normalized Error Metrics: For assessing reconstruction accuracy:
    \[
        \text{NMSE} = \frac{\|\hat{w} - w\|_2^2}{\|w\|_2^2}, \quad \text{NMAE} = \frac{\|\hat{w} - w\|_1}{\|w\|_1}
    \]
    \item Symmetric KL Divergence: Quantifies distributional differences between guided ($P$) and non-guided ($Q$) reconstructions:
    \[
        \mathcal{D}_{\text{SKL}}(P\|Q) = \frac{1}{2}[\mathcal{D}_{\text{KL}}(P\|Q) + \mathcal{D}_{\text{KL}}(Q\|P)]
    \]
    where $\mathcal{D}_{\text{KL}}(P\|Q) = \sum_{i} P(i)\log\frac{P(i)}{Q(i)}$
\end{itemize}

The Fourier space separation metrics obtained for both FLUX.1-dev and SD 2.1 base are listed in \ref{tab:watermark-dist}

\subsection{Qualitative Results}
\begin{table}[t]
\caption{Distribution Analysis of Watermarked vs Non-watermarked (Non-Attacked Images)}
\label{tab:watermark-dist}
\centering
\renewcommand{\arraystretch}{1.3}
\begin{tabular}{llcc}
\toprule
\textbf{Model} & \textbf{Image Type} & \textbf{$||\hat{w}-w||$} & \textbf{Symmetric KLD} \\
\midrule
\multirow{2}{*}{FLUX.1-dev} & Watermarked & $22.77_{0.550}$ & \multirow{2}{*}{$18.00_{0.067}$} \\
& Non-watermarked & $39.395_{1.111}$ & \\
\midrule
\multirow{2}{*}{SD 2.1 base} & Watermarked & $45.601_{1.702}$ & \multirow{2}{*}{$17.81_{0.081}$} \\
& Non-watermarked & $79.263_{2.308}$ & \\
\bottomrule
\multicolumn{4}{l}{\small Note: L1 Distance measured in Fourier space. Symmetric KLD computed between} \\
\multicolumn{4}{l}{\small watermarked and non-watermarked distributions. Subscripts denote standard deviations.}
\end{tabular}
\end{table}

To ensure experimental reproducibility, we maintained a consistent global random seed when generating initial latents across all experiments. The identical watermark key was employed throughout all tests, and we implemented a uniform timestep schedule for both sampling and inversion processes, as our preliminary tests demonstrated this approach significantly enhanced inversion quality.

\begin{figure}[!htb]
\centering
\includegraphics[width=1.15\textwidth]{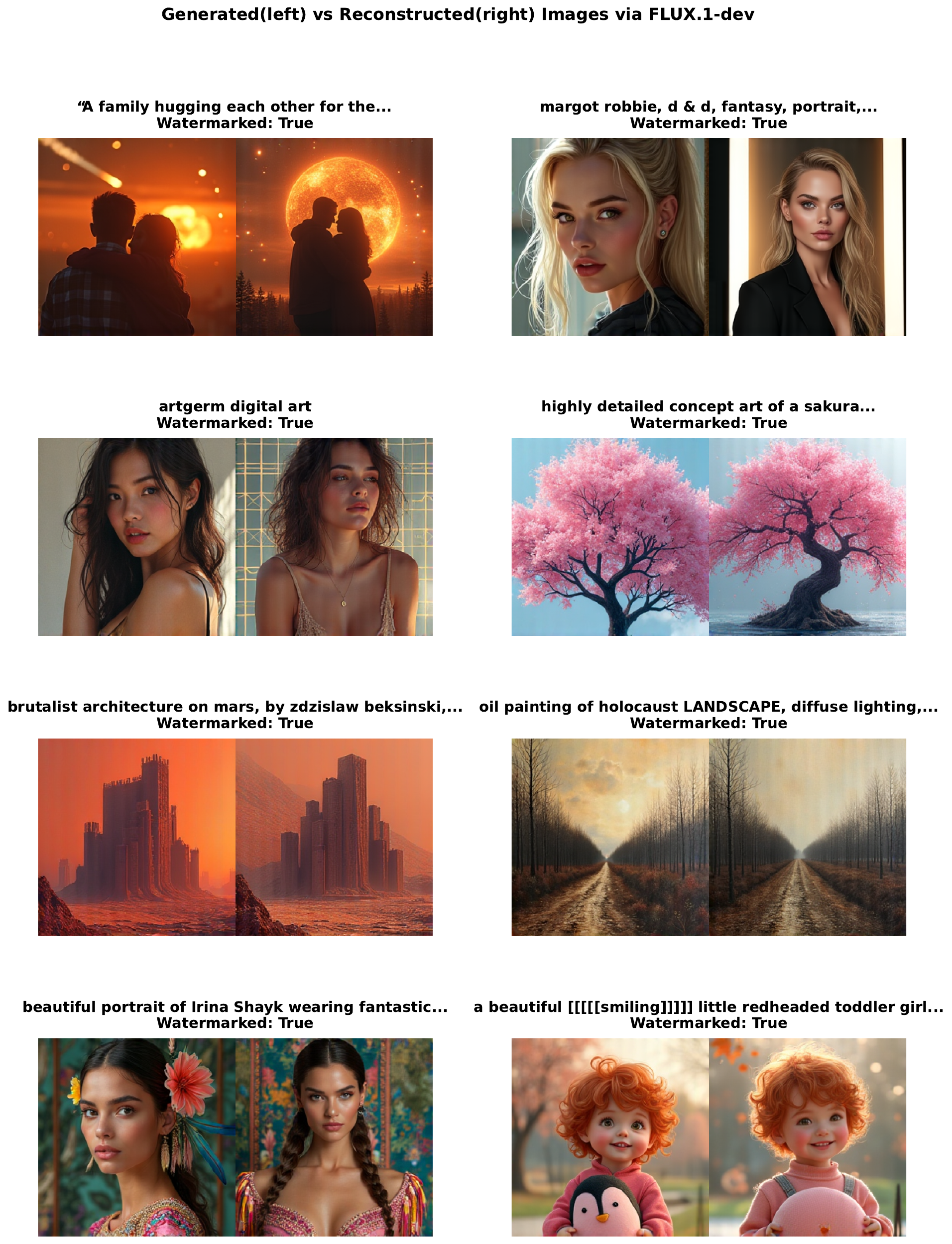}
\caption{Image generation results from the reconstructed initial noise using FLUX.1-dev. Despite using identical prompts, notable differences can be observed between original generations (left) and those from reconstructed noise (right).}
\label{qual}
\end{figure}

A notable observation from our experiments is that, even when using identical prompts, images generated from the original and reconstructed noise latents show perceptible differences, as illustrated in Figure \ref{qual}.

\begin{figure}[h!]
\centering
\includegraphics[height=2cm]{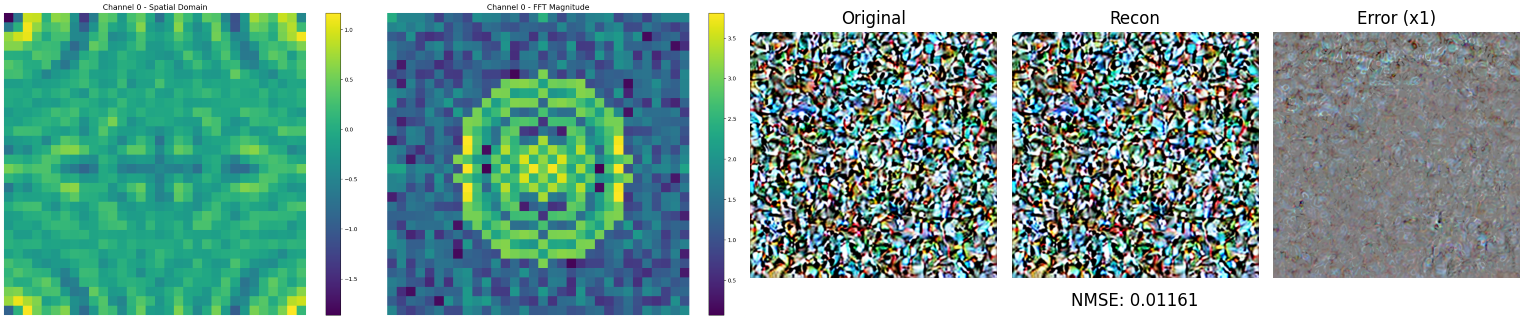}
\caption{Visualization of noise reconstruction in spatial and frequency domains. Left: Channel 0 of the latent noise in spatial domain averaged over 100 samples, showing the characteristic noise pattern. Center: Magnitude of the 2D Fourier transform of Channel 0, revealing the circular watermark pattern in frequency space. Right: Original noise, reconstructed noise, and their difference (error magnified by 1×) for a representative sample, with NMSE of 0.01161.}
\label{fig:noise_reconstruction}
\end{figure}

